\documentclass[letterpaper, 10 pt, conference]{format/ieeeconf}  

\IEEEoverridecommandlockouts                              

\usepackage{graphics} 
\usepackage{epsfig} 
\usepackage{mathptmx} 
\usepackage{amsmath} 
\usepackage{amssymb}  

\usepackage{amsthm}
\DeclareMathAlphabet{\mathcal}{OMS}{cmsy}{m}{n} 
\usepackage{amsfonts}
\DeclareMathOperator*{\argmax}{arg\,max}

\theoremstyle{definition}

\usepackage[linesnumbered,algoruled,boxed,vlined, noend]{algorithm2e}

\usepackage{hyperref}
\usepackage{tabularx}
\usepackage{epstopdf}
\usepackage{overpic}
\usepackage{xcolor} 
\usepackage{adjustbox}
\usepackage{wrapfig}
\usepackage{dsfont}

\theoremstyle{definition}
\newtheorem{definition}{Definition}

\title{\LARGE \bf
Safe and Effective Picking Paths in Clutter\\ given Discrete Distributions of Object Poses
}

\author{Rui Wang, Chaitanya Mitash, Shiyang Lu, Daniel Boehm, Kostas E. Bekris
\thanks{Work  by  the  authors  has  been  supported  by  NSF  awards  1723869, 1734492 and 1934924. The authors are affiliated with the Department of Computer Science, Rutgers University, New Brunswick, NJ, 08901, USA. Email: {\tt\small \{rw485,cm1074,sl1642\}@scarletmail.rutgers.edu, kostas.bekris@cs.rutgers.edu}.}%
}

\begin{document}

\maketitle
\thispagestyle{empty}
\pagestyle{empty}

\begin{abstract} Picking an item in the presence of other objects can be challenging as it involves occlusions and partial views. Given object models, one approach is to perform object pose estimation and use the most likely candidate pose per object to pick the target without collisions. This approach, however, ignores the uncertainty of the perception process both regarding the target's and the surrounding objects' poses. This work proposes first a perception process for 6D pose estimation, which returns a discrete distribution of object poses in a scene. Then, an open-loop planning pipeline is proposed to return safe and effective solutions for moving a robotic arm to pick, which (a) minimizes the probability of collision with the obstructing objects; and (b) maximizes the probability of reaching the target item. The planning framework models the challenge as a stochastic variant of the Minimum Constraint Removal ({\tt MCR}) problem. The effectiveness of the methodology is verified given both simulated and real data in different scenarios. The experiments demonstrate the importance of considering the uncertainty of the perception process in terms of safe execution. The results also show that the  methodology is more effective than conservative {\tt MCR} approaches, which avoid all possible object poses regardless of the reported uncertainty.
\end{abstract}

\section{Introduction}

Item picking arises in many robot manipulation applications. It involves the integration of perception and planning for recognizing the target item and then computing the motion of a robot arm for picking it. Clutter, however, can significantly complicate the challenge as it introduces occlusions and partial views.  It reduces the fidelity of object recognition and introduces uncertainty, both for the target item as well as surrounding objects. One solution - given access to an RGB-D sensor - is to compute a picking path which does not collide with the point cloud. In clutter, however, considering only the visible point cloud frequently results in collisions as it does not include the back side of objects, which may be close or attached to the target item. 

\vspace{-0.03in}
\subsection{Setup given Uncertainty in Perception}
\vspace{-0.02in}
This work focuses on the case where models of objects are available but it is unknown which of these objects are present in the scene except the target item. Given object models, object recognition and 6D pose estimation algorithms allow to identify which objects are present and their poses. Many 6D pose estimation algorithms \cite{Narayanan:2017aa, Mitash:2018aa, xiang2017posecnn} depend on point cloud registration
\cite{Mellado:2014aa}. This process first generates pose hypotheses for the object, which are then scored on how well they align with the observed point cloud.  They typically return the most likely pose hypothesis.  This process, however, ignores the effects of clutter, which introduces uncertainty in pose hypothesis generation and point cloud alignment. Ambiguities arise due to occlusions as well as imperfect learning-based prior models. For instance, when training only over simulated data, which is often a necessity as it scales
better over many object models.

\begin{figure}[t]
\centering
\includegraphics[width=\linewidth]{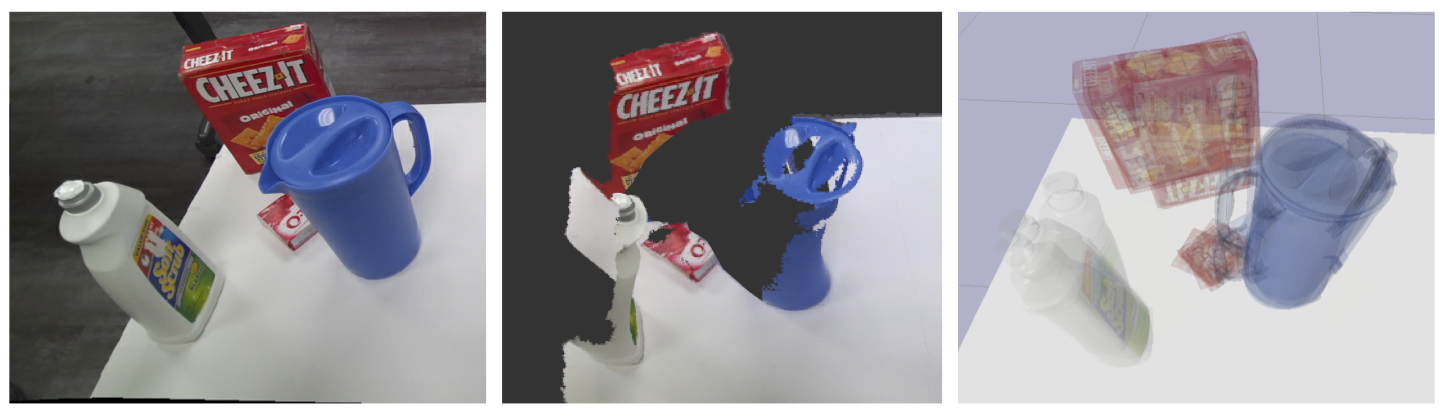}
\vspace{-.2in}
\caption{Left: An RGB image for a scenario considered in
  this work. The target ``Jell-o'' object is located between and in
  close proximity to other objects. Center: Point cloud obtained from
  the RGB-D sensor highlighting the occlusions and partial object
  views. Right: Example output of pose estimation returning a discrete
  set of poses for objects in the scene.}
\label{fig:uncertain scene}
\end{figure}

Experiments in this paper show that using only the most likely
pose estimate for each surrounding object frequently results in
collisions. The idea is to consider a discrete set of pose hypotheses
per object and define the likelihood for each hypothesis given how
well they match with the point cloud. This gives rise to a discrete
distribution of object poses. Figure \ref{fig:uncertain scene} gives a
real scene example and uses 6D pose estimation based on prior work \cite{mitash_ijrr} to define a discrete distribution of object poses.  Then, the planning problem is to compute a path of low collision risk that attaches the arm to the target given this discrete pose distribution and the likelihood of each object being in the scene.

\begin{figure*}[ht]
  \centering
  \includegraphics[width=0.95\textwidth]{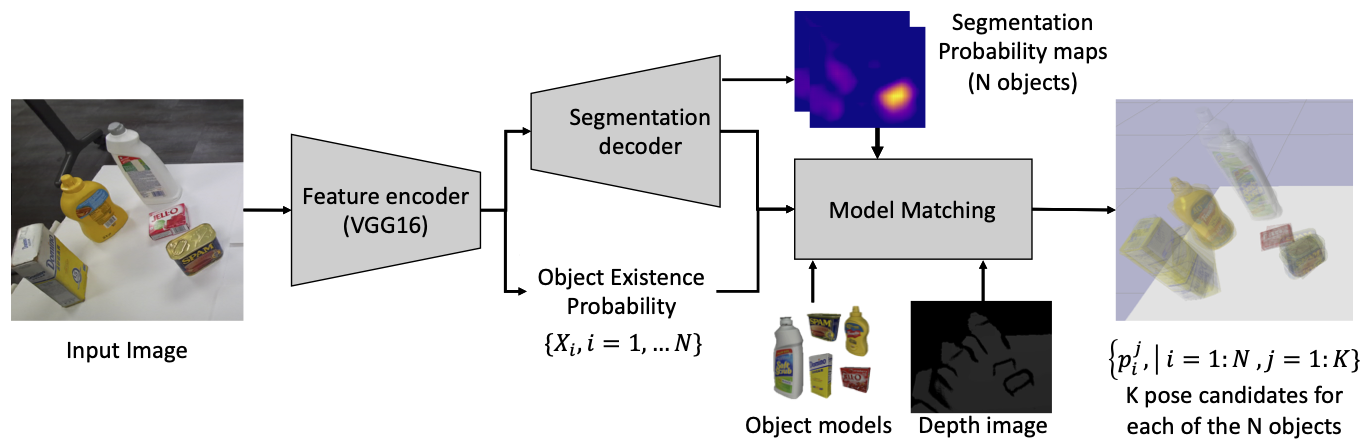}
  \vspace{-.1in}
\caption{ Perception pipeline developed to return the likelihood of each
  object existing in the scene and a discrete distribution of object
  poses.}
  \label{fig:perception_pipeline} 
\end{figure*}

\vspace{-0.01in}
\subsection{Relation to the Motion Planning Literature}
\vspace{-0.02in}

Planning under uncertainty is often formulated as a Partially
Observable Markov Decision Process (POMDP) \cite{Kaelbling:1998aa},
where a solution is a policy.  Finding an optimal policy can be
intractable \cite{papadimitriou1987complexity}, which motivates
approximations, such as computing a plausible open-loop plan and
updating it as more information is acquired
\cite{Platt:2010aa}. Updating the scene, however, can be
computationally expensive and, for the considered task, new
observations do not necessarily provide additional information. This
motivates open-loop planning for the safest path, which does not consider future observations, such as conformant probabilistic planning ({\tt CPP}) \cite{Taig:2013aa}. {\tt CPP} approaches have been used in
robotics task such as rearrangement under uncertainty
\cite{dogar2012planning, koval2015robust, anders2018reliably}.  Recent
work formalizes manipulation under uncertainty as the
blindfolded robot setting \cite{saund2019blindfolded} where the
obstacles are only sensed through contact. A pick-and-place system has been designed \cite{zeng2018robotic} to enable manipulation on novel objects with little prior knowledge. The current paper has the objective of finding safe, open-loop solutions. It focuses on discrete distributions for the presence of collision volumes and also tackles the case where the target itself is uncertain.


A heuristic way to achieve safety is to stay sufficiently away from
obstacles by growing the robot's shape by an uncertainty bound
\cite{Bry:2011aa}. If the obstacles' geometry is uncertain, the area
around estimated obstacles can be expanded into a shadow whose size
depends on uncertainty \cite{lee2013sigma}. Often the model of uncertainty assumes
a continuous Gaussian probability distribution
\cite{Park:2016aa,Sun:2016aa}.  Recent work computes obstacles geometric bounds with only partial shape information, assuming Gaussian distributed faces \cite{axelrod2018provably,
  axelrod2018hardness}. There has not been much prior work, however,
which focuses on discrete sets of object poses, which can be an
effective way to model multi-modal distributions.

\vspace{-0.03in}
\subsection{Contributions}
\vspace{-0.02in}
{\bf A.} This paper proposes a perception pipeline to predict object
existence in a scene and return the corresponding pose hypotheses with
associated probabilities on top of modern 6D pose estimation algorithms \cite{mitash_ijrr}. It integrates this perception output with a planning approach to maximize the probability of finding collision-free and successful picking paths.

{\bf B.} This paper identifies the connection between planning under
discrete models of uncertainty with failure-explanation planning problems, such as Minimum Constraint
Removal ({\tt MCR}) problems \cite{hauser2013minimum,
hauser2014minimum},
which are known to be computationally hard. While typical motion planning assumes a
collision-free solution can be found, tasks considered
here can result in no obvious collision-free solution. For instance,
if all possible object poses are avoided, then the target item is not
reachable. {\tt MCR} paths minimize the number of constraints to be removed to admit a constraint-free solution
\cite{hauser2014minimum, Erickson2013A-Simple-but-NP,
  krontiris2015computational}. The planning challenge of this work can
be seen as a stochastic version of {\tt MCR}, where instead of
minimizing the number of constraints, the
objective is to minimize the sum of the constraints' weights, i.e.,
the probabilities of the objects occupying poses along the solution
path. In this way, this work promotes the use of failure-explanation
planning in solving planning problems under uncertainty.

{\bf C.}  Experiments have been performed in simulation and with
real data in different setups (table or
shelf) and scenarios (clutter, narrow
passage, an ``arch'' of objects).  The experiments show the importance of considering multiple object poses in contrast to most
likely pose alternatives and 
the benefits of the
stochastic formulation of {\tt MCR} proposed here, which is more
effective than a conservative adaptation of {\tt MCR}, which avoids all
possible object poses.

\section{Generation of Discrete Pose Distributions}
\label{Perception pipeline}

The first task is to recognize which objects are in the scene
and their poses.  This work
utilizes the perception pipeline in
Fig.~\ref{fig:perception_pipeline}. The workspace
$\mathcal{W}$ is assumed to contain some known obstacles
$\mathcal{O}_{st}$ (e.g., a table or a shelf), and can contain any of
up to $N$ objects from a set $\mathcal{O}_{obj} = \{O_1, \dots ,
O_N\}$, for which 3D models are available.  There is a target item $O_t$ in the scene, for which a model is also available.

\vspace{-0.03in}
\subsection{Learning the existence probability of objects}
\vspace{-0.02in}
Given an RGB image, a fully convolutional neural network ({\tt
  FCN}) \cite{shelhamer2016fully} is designed to detect the objects
in the scene and to compute their segmentation
masks. The neural network comprises a {\tt VGG16} feature encoder
\cite{simonyan2014very}, followed by {\it classification} (lower
branch in Fig. 2) and {\it segmentation} (top branch in Fig. 2). The
classification outputs confidence scores
corresponding to the probability $X_i$ of each object $O_i \in
\mathcal{O}_{obj}$ detected in the scene. The segmentation outputs $N$
probability masks, one corresponding to each object possibly
in the scene. Each pixel in an objects’ probability mask indicates the
chance of the object being present at that pixel. 

\vspace{-0.03in}
\subsection{Obtaining object pose hypotheses}
\vspace{-0.02in}
Given the probability maps from the segmentation, a geometric model matching process \cite{mitash2018robust} is initiated for all objects with $X_i$ greater than a threshold (0.3 in experiments). The process samples and evaluates a large number of pose hypotheses for each object. The poses are scored based on the point cloud matching between the observed point cloud and the object model placed at hypothesized poses. The poses are then sorted based on their matching scores and clustered. The clustering  iterates over the poses in order of their scores. If a  pose hypothesis is close to a higher-ranked pose (within 2.5cm and 15 degrees), it is clustered with and represented by the higher-ranked one. This ensures that similar poses are not selected and the representative is the one with highest alignment score. Finally, the top $K$ pose representatives for each object $O_i$ are returned with scores normalized to sum up to the existence probability, $X_i$. Denote $p_i^j$ as the $j$-th pose of object $O_i$ and $Pr(p^j_i)$ as the probability that object $O_i$ will appear at pose $p^j_i$. Then:

\vspace{-.1in}
\begin{equation}
    X_i = \sum_{j=1}^{K} Pr(p_i^j) \leq 1, \ \ \ \ \ \forall i \in [1,\dots,N].
\end{equation}
\vspace{-.05in}

The target item is assumed to be in the scene, i.e., $X_t = 1$. There is uncertainty, however, regarding its poses as well, i.e., poses $p_t^j$ with probabilities $Pr(p_t^j)$ are also detected for it. The number of hypotheses $K$ for each object can vary. For simplicity, the same value is used for all objects.

\section{Problem setup and notation}
\label{Problem setup and notation}

\textbf{Path Robustness.} The robustness of a path in this paper is
defined based on two aspects:

1) \textit{Minimum collision probability with objects:}
The scene will be re-sensed and replanning is performed if a collision occurs. Thus, a path with minimum collision probability reduces overall execution time to complete a task.

2) \textit{Maximum probability of reaching the target item:} Since the
target $O_t$ also carries uncertainty, a safe path may end up having low probability to pick the object, which necessitates replanning. Therefore, maximizing the probability of reaching the target object is also important for task completion.
\vspace{-0.2in}
\begin{definition}
    \label{C-space Definition}
    ($C$-space): The configuration space ($C$-space) $\mathbb{C}$ of
    the robot arm is the set of all arm
    configurations. $\mathbb{C}_{p}$ is defined as the set of arm
    configurations, which end up in collision with an object pose $p$.
\end{definition}
\vspace{-0.15in}
\begin{definition}
    \label{Goal configurations}
    (Goal configurations): $\mathbb{Q}_{goal}$ is a set of 
    configurations where the arm can pick the target object at
    poses $p_t^j$. 
    $\mathcal{T}=[1,\dots,K]$ is the set of indices of
    all $K$ target poses. 
    Each goal configuration $q_g \in
    \mathbb{Q}_{goal}$ is associated with a set $J(q_g) \subseteq
    \mathcal{T}$ indicating which target poses $q_g$ can pick. Then,
    the probability that a path $\pi$ from the initial configuration
    $q_s$ to any goal configuration $q_g$ leads to a successful picking
    of $O_t$ is $Pr(q_g) = \sum_{j \in J(q_g)}
    Pr(P_t^j)$, i.e., equal to the probability that the target $O_t$
    is at one of the poses $q_g$ can pick.
\end{definition}
\vspace{-0.12in}
\begin{definition}
    \label{Survivability Definition}
    (Survivability of a path) A path $\pi: [0, 1] \rightarrow
    \mathbb{C}, \pi(0)=q_s, \pi(1) \in \mathbb{Q}_{goal}$
    \textit{survives} an object $O_i$, if it does not collide with
    $O_i$. The \textit{survivability} of a path $\pi$, denoted as
    $S_{\pi}$, is the probability that $\pi$ survives all the objects.
\end{definition}
    
Define $E_i$ as the event that $\pi$ collides with object $O_i$ and $\overline{E_i}$ its
complementary event. Then, $E_i^j$ is defined as the event that $\pi$
collides with object $O_i$ when $O_i$ is at pose $p_i^j$.  Then $E_i =
\bigcup_{j=1}^{K} E_i^j$. The events that $O_i$ appears at different candidate poses are mutually exclusive, which indicates mutual exclusiveness of $E_i^j$. Thus, the probability that $\pi$ does not collide with $O_i$
is:
\begin{equation}
    Pr(\overline{E_i}) = 1 - Pr(E_i)  
    = 1 - Pr(\bigcup_{j=1}^{K} E_i^j) = 1 - \sum_{j=1}^{K} Pr(E_i^j).
\end{equation}
Then, the survivability of a path for all objects $O_i \in
\mathcal{O}_{obj} (i=1,\dots,N)$ is computed as:
\begin{equation}
    \label{survivability of path}
        S_{\pi} = Pr(\bigcap_{i=1}^N \overline{E_i}) = 
        \prod_{i=1}^N Pr(\overline{E_i}) = 
        \prod_{i=1}^N (1 - \sum_{j=1}^{K} Pr(E_i^j)).
\end{equation}

$S_{\pi}$ represents the first aspect of path robustness, i.e., minimum collision probability with objects. The higher $S_{\pi}$ is, the less risky the path is in terms of collision.

\begin{definition}
    \label{Success probability value}
    (Success probability): Define $E$ as the event that the path
    survives all objects and $F$ as the event that the robot arm
    reaches the target item. Both events E and F must occur for a path
    $\pi: q_s \rightarrow q_g$ to successfully reach the target.
    Define the success probability of a path to be $Succ(\pi) = Pr(E,
    F)$.  Then, $Succ(\pi)$ can be rewritten as:
    \begin{equation}
        Succ(\pi) = Pr(E, F) = Pr(E) \cdot Pr(F) = S_{\pi} \cdot Pr(q_{g}|\pi).
    \end{equation}
\end{definition}

The events $E$ and $F$ are
independent with one exception, which leads to
defining $Pr(F)=Pr(q_g|\pi)$.  When a path leading to a goal $q_g$
collides with the target pose $p_t^j$ with which $q_g$ is associated, $q_g$ is no longer considered for the path as a valid
goal configuration for $p_t^j$, since: (1) if the target item $O_t$ is
at pose $p_t^j$, the path will be in collision with $O_t$ or (2) if
the target item $O_t$ is not at pose $p_t^j$, then $q_g$ does not
allow picking $O_t$ at pose $p_t^j$.

This means that $Pr(q_{g})$ should be conditioned on the path. Define
${\overline{J}}_{\pi}$ as the target poses intersected by the
path. Then, the remaining valid target poses for $q_g$ should be
$\hat{J}_{\pi}(q_g) = J(q_g) \setminus {\overline{J}}_{\pi}$ and
$Pr(q_g|\pi)$ is set to be:

\vspace{-.15in}
\begin{equation}
    \label{conditional definition of Pr(q_{g})}
    Pr(q_g|\pi) = \sum_{j \in \hat{J}_{\pi}(q_g)} Pr(P_t^j).
\end{equation}

Define the path space $\Pi$, which includes the set of candidate paths
$\pi: q_s \rightarrow q_g$, where $q_g \in \mathbb{Q}_{goal})$. The
overall objective is to find a path $\pi^*$:

\vspace{-.15in}
\begin{equation}
    \pi^* = \argmax_{\pi \in \Pi} Succ(\pi).
\label{eq:max_success}
\end{equation}

\vspace{0.1in}

\section{Algorithmic Framework}
\label{Algorithmic Framework}

Consider a roadmap $G(V, E)$, where $q \in V$ corresponds to an arm configuration and $e \in E$ the transition between two arm configurations.
If an edge $e$
connecting $q_1$ and $q_2$ intersects with a pose $p^j_i$, a label $l^j_i$  is
assigned to that edge and the weight for the label $w(l^j_i) = Pr(p^j_i)$.

The survivability $S_{\pi}$ depends on the probability
that the objects will appear at any pose that the path
intersects.  So $Pr(E^j_i)$ in Eq.\ref{survivability of path}
depends on whether there is an edge $e$ along the path $\pi$ that has
a label $l_i^j$. If such an edge exists along the path, then
$Pr(E_i^j)=Pr(p_i^j)=w(l_i^j)$.  Otherwise, $Pr(E_i^j)=0$.  An
indicator random variable $\mathds{1}_{\pi}(j,i)$ for each pose $j$
of each object $i$ is defined as
\begin{equation}
    \label{indicator random variable}
        \mathds{1}_{\pi}(j,i) =
        \begin{cases}
            1, & \text{if}\ \pi \ \text{carries label}\  l_i^j \\
            0, & \text{otherwise}.
        \end{cases} 
\end{equation}
Then $S_{\pi}$ in a labeled roadmap can be
computed as
\begin{equation}
    \label{survivability in labeled roadmap}
    S_{\pi} 
    = \prod_{i=1}^N (1 - \sum_{j=1}^{K} Pr(E_i^j))
    = \prod_{i=1}^N (1 - \sum_{j=1}^{K} w(l_i^j)\mathds{1}_{\pi}(j,i)).
\end{equation}

Eq. \ref{survivability in labeled roadmap} computes
$S_{\pi}$ for a path from $q_s$ to any currently examined
state $q_{curr}$ by checking the labels $L_{q_{curr}}$ the path $\pi:
q_s \rightarrow q_{curr}$ carries.  To compute the prospect of the
path $\pi$ accurately reaching the target, two situations are
considered:

{\em (i)} If $q_{curr}$ is a goal configuration $q_g$, the
probability that the path leads to the goal is computed according
to Eq. \ref{conditional definition of Pr(q_{g})}.

\begin{figure}[h]
  \begin{center}
    \includegraphics[width=0.25\textwidth]{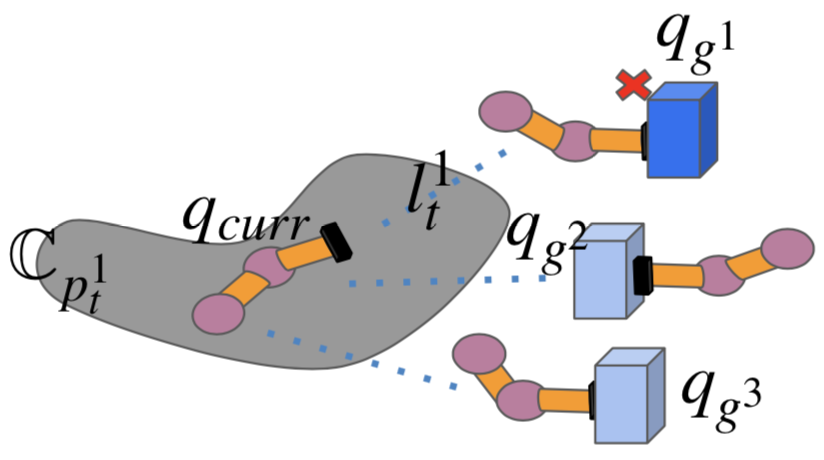}
    \includegraphics[width=0.21\textwidth]{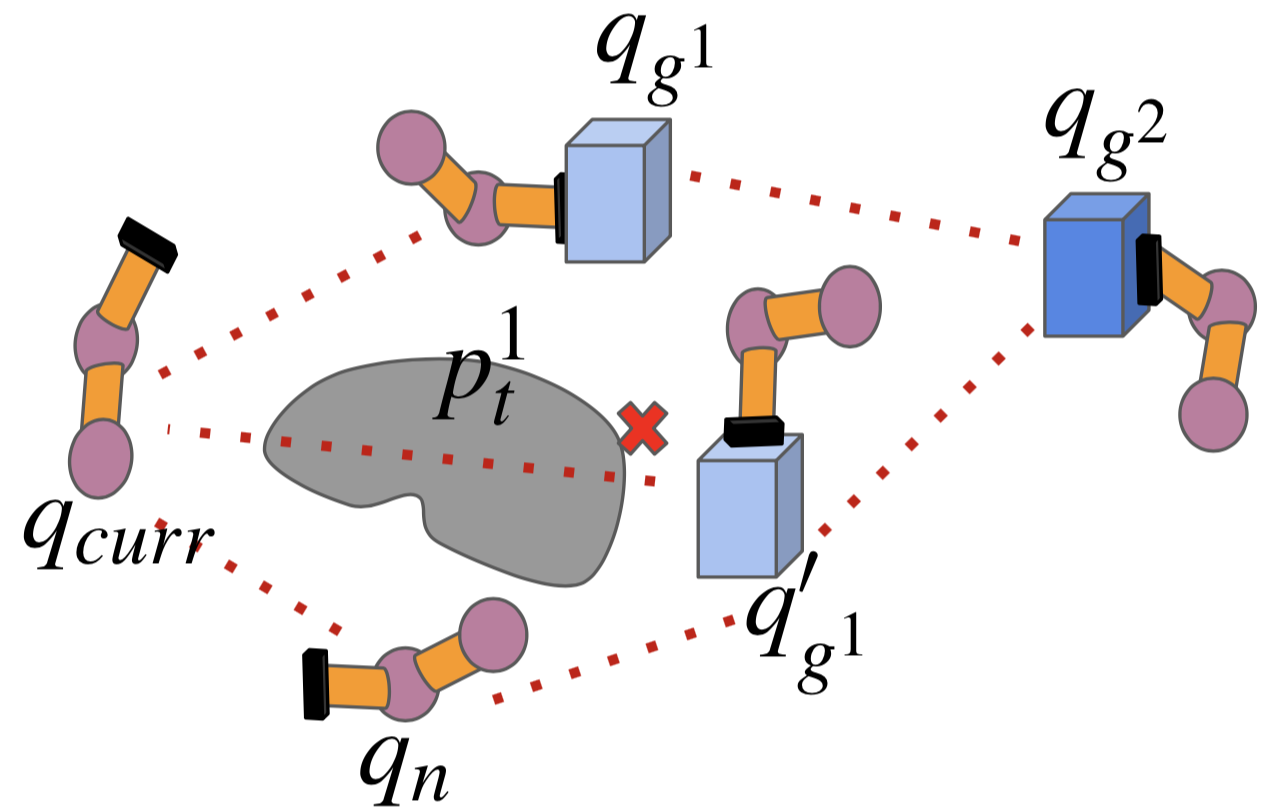}
  \end{center}
  \vspace{-.1in}
  \caption{(Left) A path $\pi: q_s \rightarrow q_{curr}$ intersects pose $p_t^1$, which makes $q_{g^1}$ no longer valid for this path. 
    The remaining goals will be $\{q_{g^2}, q_{g^3}\}$ and $T_{q_{curr}} = [2,3]$ 
    (Right)
    The goal $q^\prime_{g^1}$ is not
    available since the path from $q_{curr}$ to $q^\prime_{g^1}$
    collides with pose $p_t^1$.  But it is still  available if being reached via $q_n$.  $q_{g^1}$ can also be treated as an intermediate
    configuration to reach a potentially more promising goal
    $q_{g^2}$. The child node for the path ending at $q_{g^1}$ will be
    added to search twice, once as a goal node, once as a non-goal node. \vspace{.1in}}
\label{fig:alpha}
\end{figure}

{\em (ii)} If $q_{curr} \notin {Q}_{goal}$, the
path is not complete and $T_{q_{curr}} \subseteq \mathcal{T}$ indicates the indices of remaining target poses
the path $\pi:
q_s \rightarrow q_{curr}$ can reach. 
If the current path $\pi:
q_s \rightarrow q_{curr}$ carries a label $l_t^j$, 
then the path and its extensions can no longer treat $p_t^j$ as a valid target pose. In this case, $T_{q_{curr}}$ is updated by removing pose index $j$ from $T_{q_{curr}}$.
An example is given in Fig. \ref{fig:alpha}(left).
Then the probability for the current path $\pi: q_s \rightarrow q_{curr} $ leading to the true target will be
\vspace{-.01in}
\begin{equation}
    \label{reaching_probability}
    Pr(q_{curr}|\pi) =  \sum_{j \in (T_{q_{curr}}\setminus{\overline{J}}_{\pi})} Pr(p_t^j)
\end{equation}

With Eq. \ref{conditional definition of Pr(q_{g})}, \ref{survivability in labeled roadmap} and \ref{reaching_probability}, the success probability $Succ(\pi)$ of a path $\pi : q_s
\rightarrow q_{curr}$ can be computed, which is used as the objective function during the search.

\subsection{Challenge - Lack of Optimal Substructure}

\begin{figure}[h]
  \begin{center}
    \includegraphics[width=0.48\textwidth]{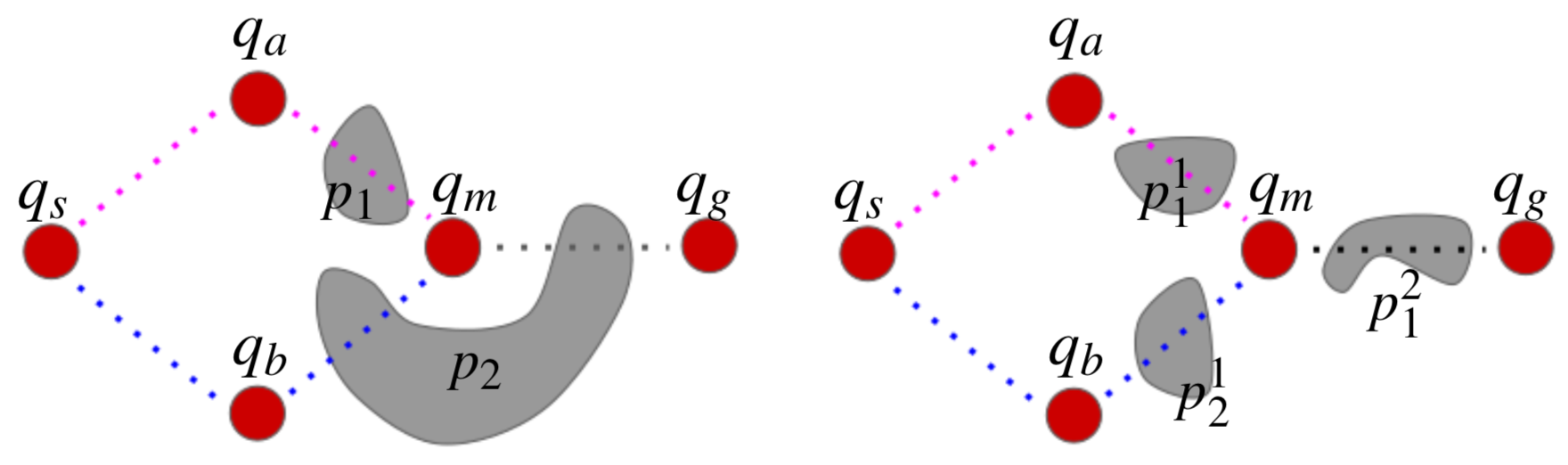}
  \end{center}
  \vspace{-.1in}
  \caption{(Left) There are 2 poses 
  $Pr(p_1)=0.3$ and $Pr(p_2)=0.4$. Two paths are considered: the
  pink path $q_s \rightarrow q_a \rightarrow q_m$ is favored vs. the blue one
  $q_s \rightarrow q_b \rightarrow q_m$ locally as it has a lower
  collision probability. Nevertheless, both paths go through
  pose $p_2$ afterwards ($q_m \rightarrow q_g$). Thus, the optimal path to $q_g$ is actually the
  blue one, which only collides with pose $p_2$ with probability 0.4.
  (Right) There are 3 poses: $p_1^1$ and $p_1^2$ for object $O_1$ with
  $Pr(p_1^1)=0.3$ and $Pr(p_1^2)=0.3$, while $p_2^1$ belongs to $O_2$
  with $Pr(p_2^1)=0.4$. Again the pink path $q_s \rightarrow q_a
  \rightarrow q_m$ is locally favored since it has higher
  survivability $1-Pr(p_1^1)=0.7$ than that of the blue path $q_s \rightarrow
  q_b \rightarrow q_m$ ($1-Pr(p_2^1)=0.6$).  But when both paths reach
  $q_g$, the pink path has survivability $1-Pr(p_1^1)-Pr(p_1^2)=0.4$,
  which is lower than that of the blue path $(1-Pr(p_2^1))(1-Pr(p_1^2))=0.42$.}
\label{fig:beta}
\end{figure}

Consider the setup in
Fig. \ref{fig:beta}(left) where the locally optimal path is not globally optimal, i.e., the search algorithm cannot just remember  the locally optimal path (greedy search). For the related {\tt MCR} problem,  it has been argued that greedy search (where only the best local path is stored at each node) still guarantees an optimal solution if the optimal path encounters each obstacle once \cite{hauser2014minimum}. Even with this assumption for the current problem, greedy search still has a chance of failing to find the optimum (Fig. \ref{fig:beta}(right)). A straightforward reduction of the problem to the computationally hard {\tt MCR} problem can be found in the Appendix.

\subsection{MaxSuccess Search}
Since greedy search is not guaranteed to be optimal,  a complete search method is proposed here, which follows the exact search for the {\tt MCR}
problem but adapts it to address the survivability objective
defined here. The algorithm stores a path to a
node only if the label set of the path is not a superset of that of
any path found reaching the same node. It theoretically guarantees optimality
since a path with a superset of labels cannot have a higher
$Succ(\pi)$ value than that of a path reaching the same node
with a subset of labels.

\begin{algorithm}
\label{alg:MSE Algorithm}
    \DontPrintSemicolon
    \KwIn{$G(V,E), q_s, \mathbb{Q}_{goal}, \mathcal{T}=[1,\dots,K]$}
    \KwOut{$\pi^*$}
        $ Q \leftarrow \text{{\sc Add}}(q_s, L_{q_s}=\emptyset, T_{q_s} = \mathcal{T}, \mathds{1}_{q_s}=false) $\;
        \While{$\textit{goal not found}$}{
            $q_{curr} \gets Q.top()$\;
            \If{$\mathds{1}_{q_{curr}} = true$}{
                \KwRet{$\pi: q_s \rightarrow q_{curr}$}
            }
            \For{$\text{each } q_{neigh} \in \text{Adj}(G, q_{curr})$}{
                $L_{q_{neigh}} = L_{q_{curr}} \cup L_e(q_{curr}, q_{neigh})$\;
                \If{$ \text{not } \textsc{IsSuperSet}(L_{q_{neigh}}) $}{
                    $q_{neigh}.{\pi} \gets q_{curr}.{\pi} \cup e(q_{neigh}, q_{curr})$\;
                    $S_{q_{neigh}.{\pi}} \gets \textsc{GetSurvival}(L_{q_{neigh}})$\;
                    $T_{q_{neigh}} \gets \textsc{UpdateGoals}(L_{q_{neigh}})$\;
                    $Pr(q_{neigh}|\pi) \gets \textsc{GetReach}(T_{q_{neigh}})$\;
                    $Succ(q_{neigh}.{\pi}) \gets S_{q_{neigh}.{\pi}} \cdot Pr(q_{neigh}|\pi)$\;
                    $Q \gets \text{{\sc Add}}(q_{neigh},L_{q_{neigh}}, T_{q_{neigh}},false)$\;
                    \If{$q_{neigh} \in \mathbb{Q}_{goal}$}{
                        \If{$\text{{\sc IsValid}}(q_{neigh}, T_{q_{neigh}})$}{
                        $Q\gets\textsc{Add}(q_{neigh},L_{q_{neigh}},T_{q_{neigh}},true)$
                        }
                    }
                }
            }
}
\caption{MaxSuccess Exact Search}    
\end{algorithm}

\begin{figure*}[t]
  \centering
  \includegraphics[width = 0.9 \textwidth]{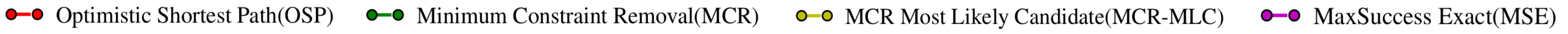}
  \begin{tabular}{cccc}
      \includegraphics[width = 0.23 \textwidth]{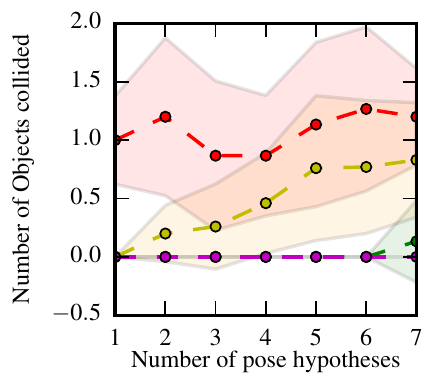} &
      \includegraphics[width = 0.23 \textwidth]{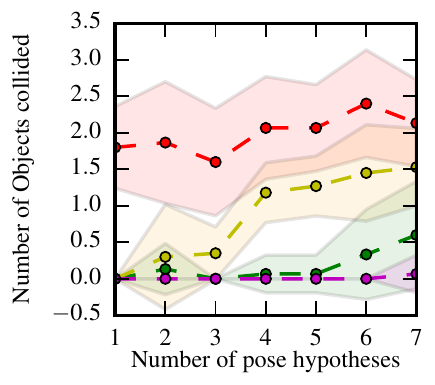} &
      \includegraphics[width = 0.23 \textwidth]{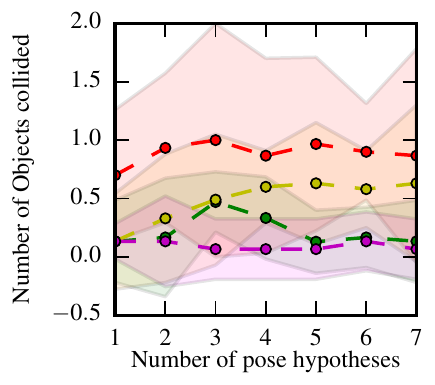} &
      \includegraphics[width = 0.23 \textwidth]{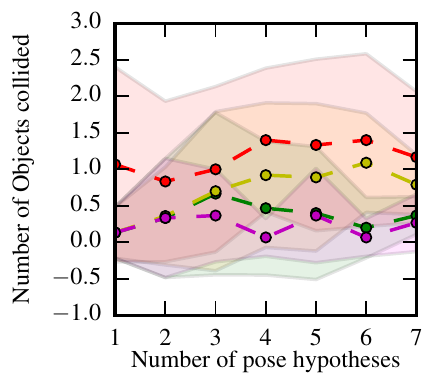} \\
      \includegraphics[width = 0.23 \textwidth]{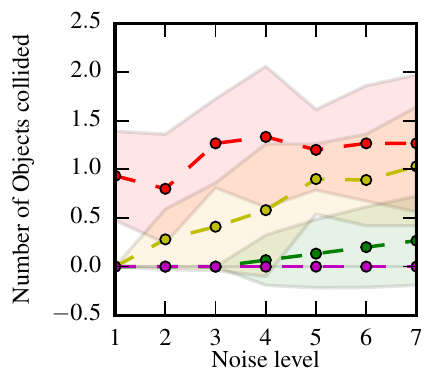} &
      \includegraphics[width = 0.23 \textwidth]{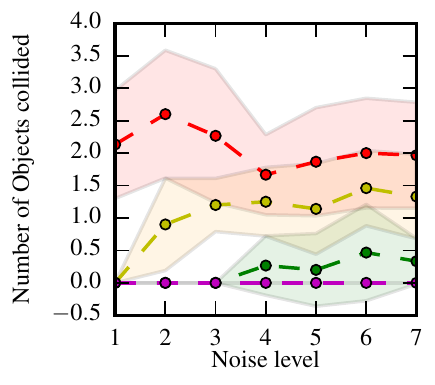} &
      \includegraphics[width = 0.23 \textwidth]{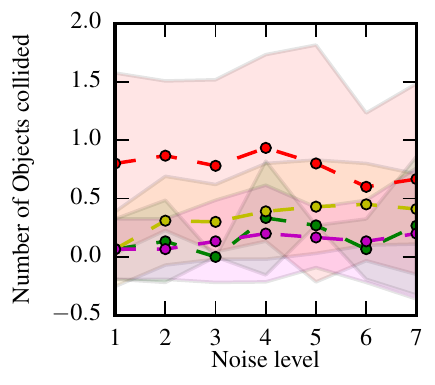} &
      \includegraphics[width = 0.23 \textwidth]{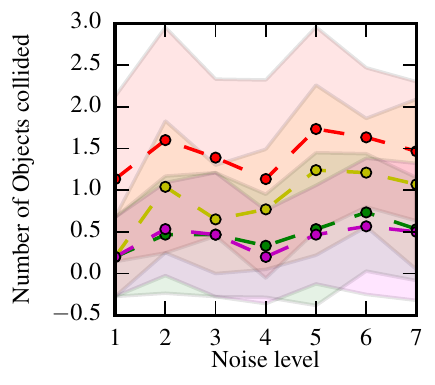} \\
      \includegraphics[height=1.2in, width = 0.23 \textwidth]{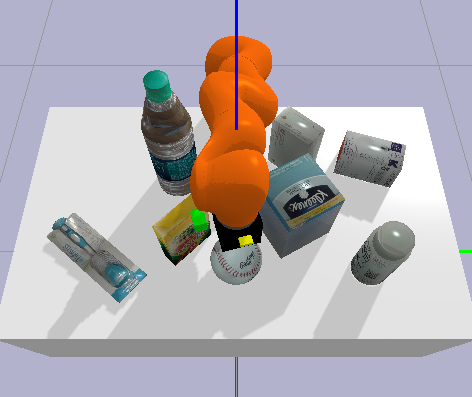} &
      \includegraphics[height=1.2in, width = 0.23 \textwidth]{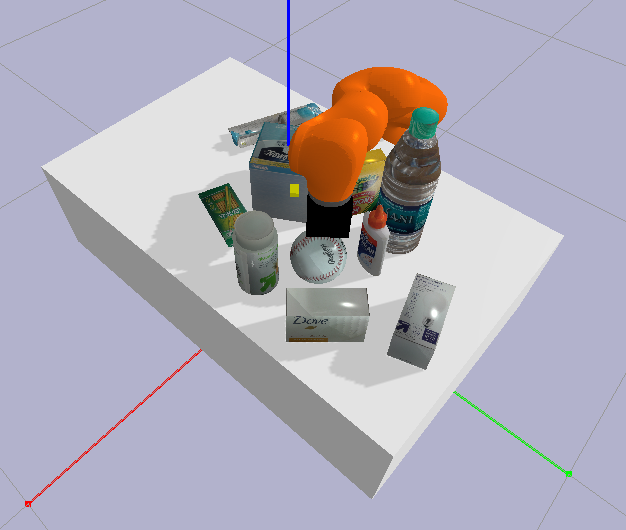} &
      \includegraphics[height=1.2in, width = 0.23 \textwidth]{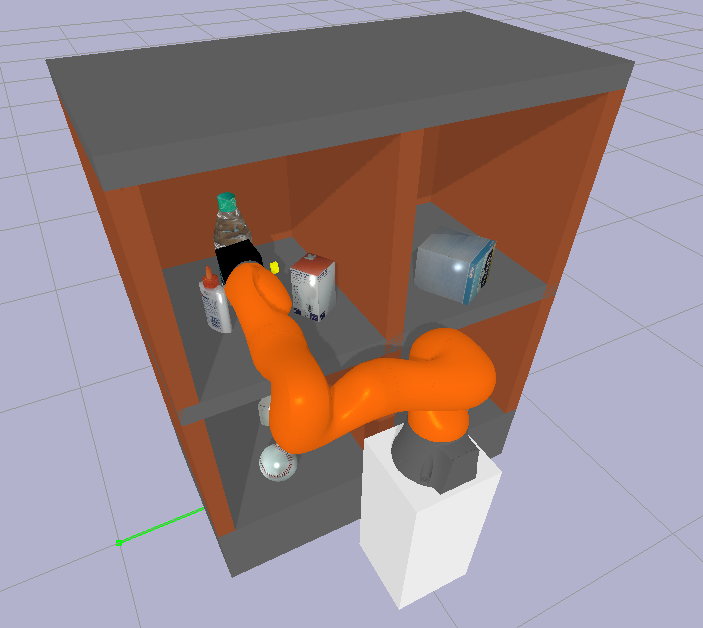} &
      \includegraphics[height=1.2in, width = 0.23 \textwidth]{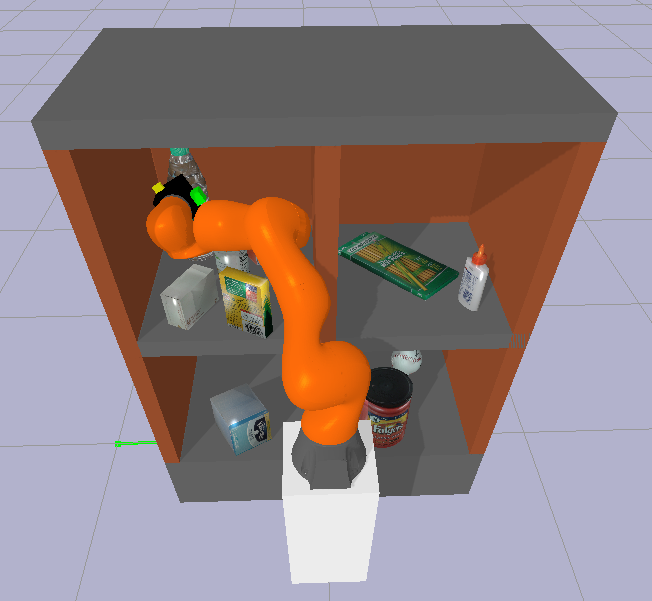} \\  
      \small Table 1 (narrow passage) & \small Table 2 (clutter) & \small Shelf 1 (narrow passage) & \small Shelf 2 (clutter) \\    
  \end{tabular}
  \caption{Results on 4 tests (2 for tabletop and 2 for
    shelf) evaluating the number of objects collided for (top) different number of pose hypotheses    under uncertainty level 4 and (middle) different uncertainty
    levels for 4 poses per object. Each column corresponds to the
    benchmark shown below it. The target object is the baseball in
    the tabletop and the water bottle in the shelf. The dots are average values and the background color indicates variance.}
 \label{fig:experimental scenarios}
\end{figure*}

The method is outlined in Alg. \ref{alg:MSE Algorithm}.  It receives as
input the roadmap $G(V,E)$, the start arm configuration $q_s$, a
set of goal configurations $\mathbb{Q}_{goal}$ and a list of all target poses indices $\mathcal{T}$.  A priority queue $Q$ (line 1) prioritizes
nodes with higher $Succ(\pi: q_s \rightarrow q)$.  Each
state $q$ is specified with a label set $L_q$ indicating the labels
the path $\pi: q_s \rightarrow q$ carries and the set of indices of the target poses $T_q \subseteq \mathcal{T}$ the path $\pi$ can
reach. 
An indicator $\mathds{1}_q$ is assigned to a node to indicate
whether it is a goal. If it is a goal, then it is found with the highest
$Succ()$ (line 4-5).  If not, it computes
the labels that the path from $q_s$ to all adjacent nodes $q_{neigh}$ via $q_{curr}$
carries (line 6-7).
If the set of labels is not
a superset of that of any previously stored paths $\pi: q_s
\rightarrow q_{neigh}$ (line 8), the path $\pi: q_s \rightarrow
q_{curr} \rightarrow q_{neigh}$ is stored, with corresponding $S_{\pi}$ and $T_{q_{neigh}}$ computed (line 9-11).
The probability for the path to reach the target is computed in line 12 using Eq. \ref{reaching_probability} and $Succ()$ value of the path can be computed (line 13).
The node is then added to
$Q$ as a non-goal node ($\mathds{1}_q=false$) (line 14). Then the algorithm
checks if it is a goal node and whether the path $\pi:q_s
\rightarrow q_{curr} \rightarrow q_{neigh}$ still treats this goal
$q_{neigh}$ as a valid one (line 15-16).  If $q_{curr}$ meets both
conditions, it is added to $Q$ as a goal
($\mathds{1}_q=true$) (line 17).
The search terminates when a goal $q_g \in \mathbb{Q}_{goal}$ has
been found (line 5) or there is no solution.

A key observation is that each node has a chance to be added to $Q$
twice in a single iteration (line 14 and line 17). If it is a non-goal
node, it will only be added once (line 14).  A goal node can be
treated as a goal configuration to grasp an object, but it can also be treated as
an intermediate node to reach a more promising goal configuration. For
that reason, the node is added as a non-goal node (line 14) as
usual. Then, the node is only added as a goal node (line 17) if it is still valid
(Fig. \ref{fig:alpha}(right)).

\vspace*{-.01in}

\section{Experimental Evaluation}
\label{Experimental Evaluation}
The proposed planning framework in \ref{Algorithmic Framework} is evaluated on (A) simulated sensing data and (B) data from a real-world setup. The following alternatives are also considered:
\begin{enumerate}
\item an $\tt Optimistic\ Shortest\ Path$ ($\tt OSP$) planner -
  which ignores the presence of the movable objects
  $\mathcal{O}_{obj}$,
\item an $\tt MCR$ search (Exact and Greedy) - which aims to minimize the
  number of collisions with all poses, and
\item an $\tt MCR \ Most\ Likely\ Candidate\ (MCR-MLC)$ search, which
  considers only the most likely pose for each object and aims to
  minimize the number of collisions.
\end{enumerate}

The methods are evaluated on two robot manipulators (1) a 7-DoF Kuka LBR iiwa14 and (2) a 7-DoF Yaskawa Motoman SDA10F, each of which with a suction-based gripper. The evaluation metrics used here are 
(1) the number of objects colliding in the ground truth scene and 
(2) 
success rate in reaching the true target.
\vspace{-0.05in}
\subsection{Simulations}
\vspace*{-0.03in}
Large-scale experiments with simulated sensing data are performed first to evaluate the algorithms with different levels of uncertainty. Four benchmarks are created (Fig. \ref{fig:experimental scenarios} bottom), where
benchmarks 1-2 are tabletop scenarios while benchmarks 3-4 are highly-constrained shelf scenarios. The target object is either placed in a narrow passage or in clutter.

\begin{figure}[t]
    \centering \includegraphics[width = 0.45
      \textwidth, height = 0.13 \textheight]{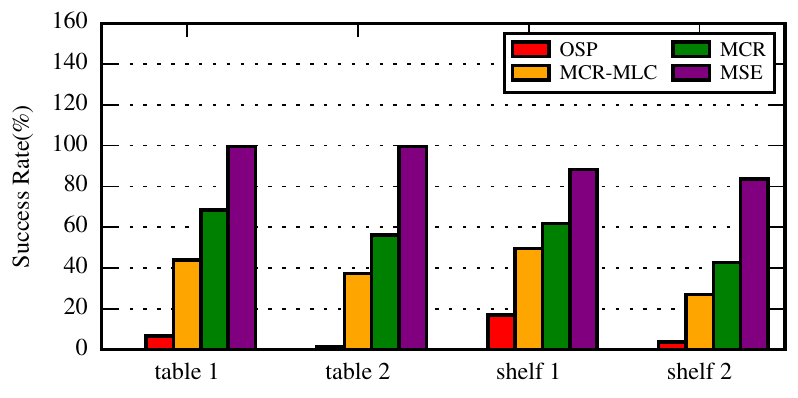}
    \vspace{-.1in}
\caption{\footnotesize 
    \label{fig:success rate} 
    Success rate for different algorithms in all 4 benchmarks.}
\end{figure}

\begin{figure*}[t]
  \centering
  \includegraphics[width=0.9\textwidth, height=0.35\textheight]{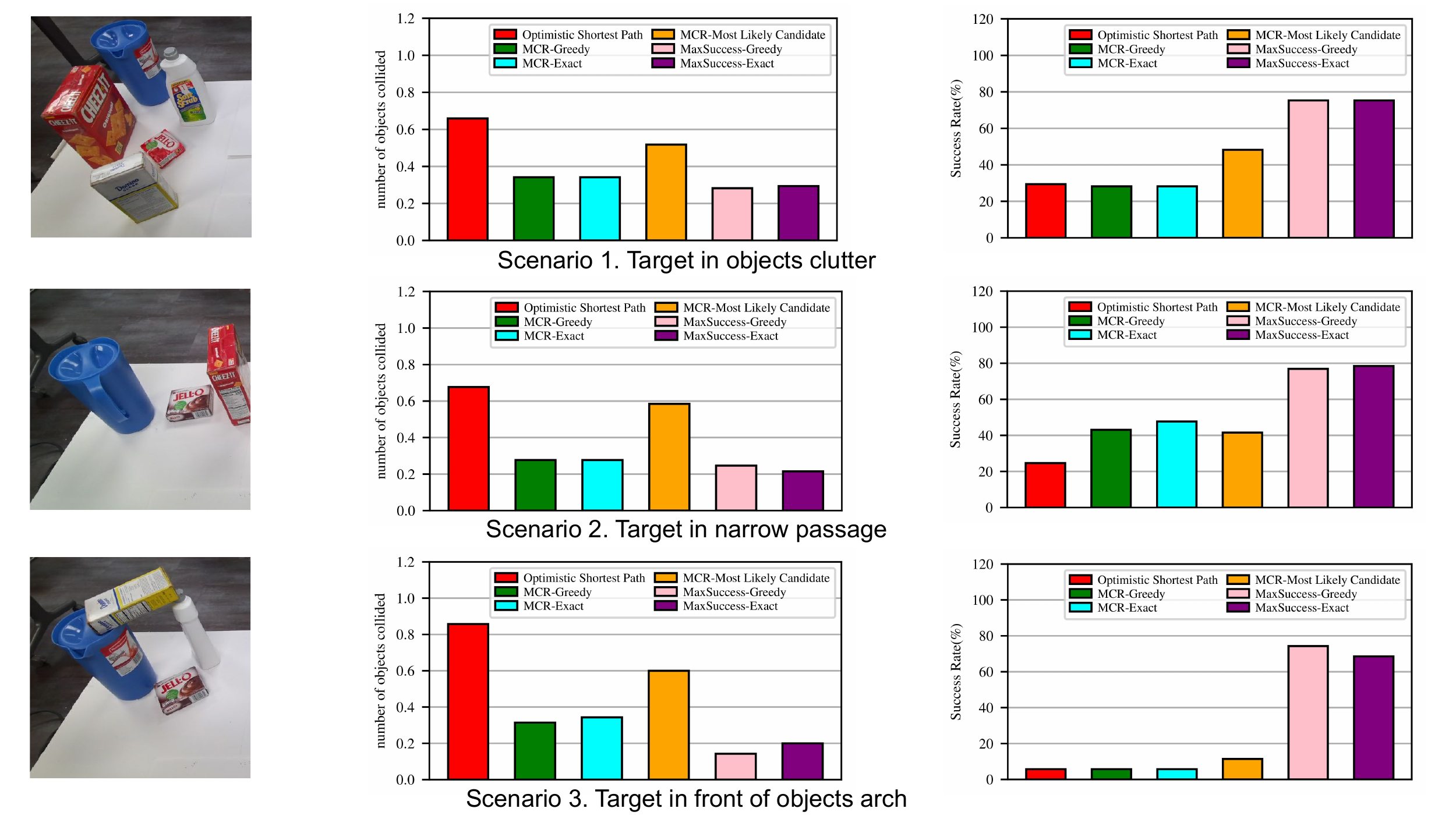}
  \vspace{-.1in}
  \caption{Experimental results on real vision system in 3 scenarios (1) target in objects clutter (top row) (2) target in narrow passage (middle row) (3) target in front of objects arch (bottom row). The second and third column demonstrate the number of objects collided during path execution and the success rate of reaching the target, respectively, as the evaluation metrics for 6 methods (including both exact and greedy version of {\tt MaxSuccess} and {\tt MCR}).}
 \label{fig:experiment on real vision system}
   \vspace{.05in}
\end{figure*}

Pose
hypotheses (sampled between 1-7) are generated according to probability distributions
centered at the ground truth pose. Different levels of uncertainty are defined (Level 1: $\pm0.5$
cm for translation error and $\pm5$ degrees for orientation error; level 7: $\pm3.5$ cm and $\pm35$ degrees noise). Intermediate levels (2-6) are linearly interpolated between
levels 1 and 7.  The pose probability is assigned based on its
distance from ground truth. 35 roadmaps are generated for each ground truth using a Probabilistic Roadmap ({\tt PRM}) \cite{kavraki1996probabilistic}. The details of the roadmap generation are provided in the Appendix. Fig. \ref{fig:experimental scenarios} provides the number of objects collided during path execution under different uncertainty.

In the tabletop benchmarks, the $\tt OSP$ paths collide with $1.54$
objects on average and $\tt MCR-MLC$ with $0.74$. In contrast, both $\tt MCR$ and $\tt MSE$ work well with much fewer collisions.  As the
uncertainty increases, the number of collisions for ${\tt
  MCR}$ methods start to increase, while the $\tt MSE$ algorithm
remains almost down to zero collisions.  Overall, $\tt MCR$ and $\tt
MSE$ result in few collisions but
Fig. \ref{fig:success rate} shows that $\tt
MCR$'s success rate is not as high (68.4\% for table 1 and 56.2\% for table 2), since it does not reason about target uncertainty. As a
result, $\tt MCR$ may avoid collisions but does not
lead to the true target.  Since the $\tt MSE$ methods take both safety
and goal reachability to form the success function $Succ(\pi)$, the
corresponding success rate is high on the tabletop scenes ($99.5\%$).

The shelf benchmarks are more challenging, as the objects must be
reached from the side in a limited space, which increases the collision risk.
Despite that, $\tt MCR$ and $\tt MSE$ remain safe
($0.34$ and $0.22$ collisions, respectively). Again, the $\tt MCR$ method is conservative
in terms of collisions at the expense of failing to reach the true
target pose.  In Fig. \ref{fig:success rate}, the success rate for
$\tt MCR$ drops to ($42.6\%$) in benchmark 4 (clutter, shelf), while
$\tt MSE$ is still able to succeed $83.6\%$ of the trials.

\vspace{-0.06in}
\subsection{Real-world Experiments}
\vspace{-0.03in}

After verifying the effectiveness of the proposed $\tt MaxSuccess $ algorithm with simulated sensing data, an evaluation with real data took place. Fig. \ref{real robot setup} shows the sensing setup used. An Azure Kinetic camera is mounted on top of a humanoid Motoman SDA10F robot to enable an overhead view of the objects on the table. The experiment focuses on overhand picks in a tabletop setup. 33 images have been taken from the camera with diverse scenarios:

\begin{wrapfigure}{R}{0.24\textwidth}
\vspace{-.1in}
    \begin{center}
        \includegraphics[width=0.24\textwidth, height=0.16\textheight]{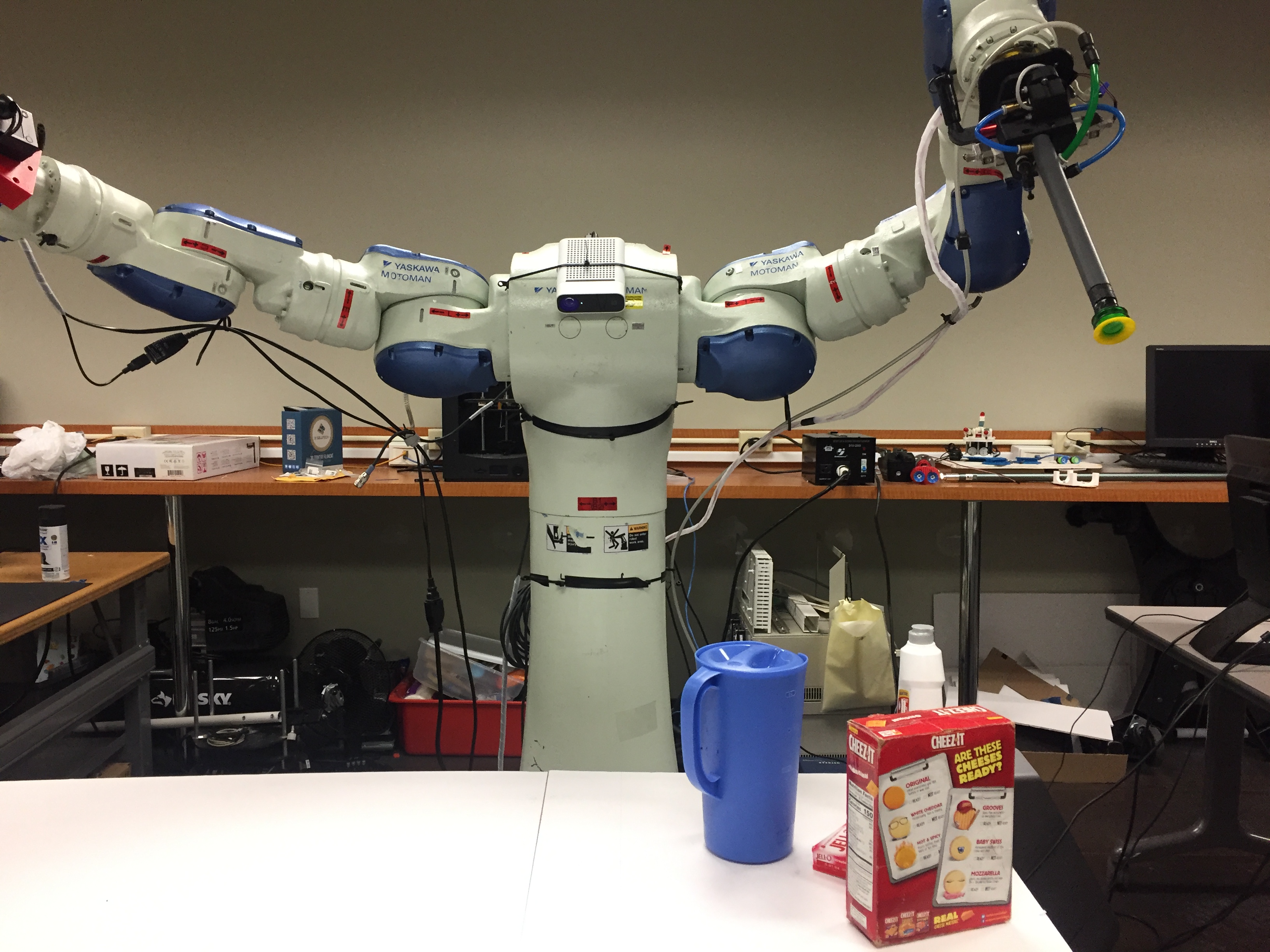}
\vspace{-.25in}
        \caption{Real system setup}
\vspace{-.1in}
        \label{real robot setup}
    \end{center}

\end{wrapfigure}

\noindent \textbf{(1) target in clutter} - Target surrounded by multiple objects. The robot has to reason about the objects' uncertainty to reach the target without collision. \textbf{(2) target in narrow passage} - Target placed between 2 or 3 tall objects to create a narrow passage. The arm has to reach deep to pick the target. \textbf{(3) target under an obstacle arch} - An arch is created by three objects, where an object is put on top of the other two. The target is placed a little bit ahead of the arch. The robot is reaching the target with overhand picks, so the relative location of the arch to the target has to be carefully examined to succeed. 

10 YCB objects are selected and the "pudding box", "gelatin box" and "meat can" are selected as the target in different scenes. All  images went through the perception pipeline as in Section \ref{Perception pipeline}. An object is treated as present if the probability prediction is over 0.3. Given this threshold, the accuracy of object recognition is analyzed in the Appendix. For each detected object, $K=5$ object poses have been returned with corresponding probabilities as the outcome of pose estimation and pose clustering\footnote{All the data of the 6D pose hypotheses can be found at \url{https://robotics.cs.rutgers.edu/pracsys/projects/planning_under_discrete_uncertainty/}}. 

The proposed planner takes these poses as input. The process generates 5 roadmaps for each scene to produce the picking paths for each method. The paths are then executed in simulation using ground truth poses to evaluate their performance. The same metrics are used: (1) number of colliding objects; and (2) success rate of reaching the target.

\begin{figure}[t]
  \centering
  \includegraphics[height=1.025in]{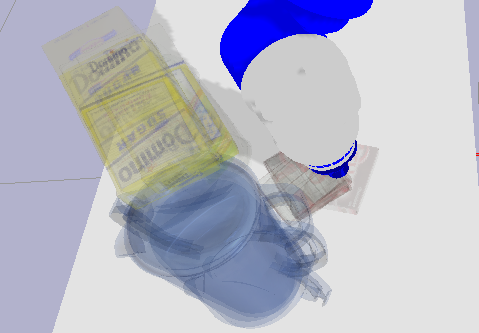}
  \includegraphics[height=1.025in]{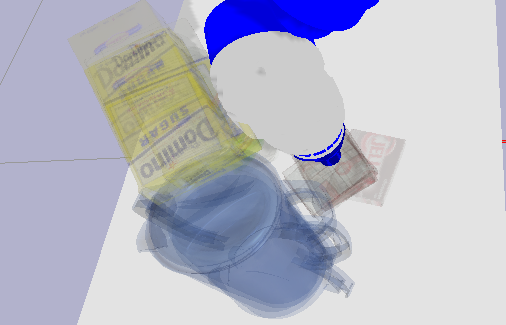}
  \vspace{-.05in}
\caption{\footnotesize 
    \label{fig:MCR vs MaxSuccess} 
Left image shows the grasping configuration chosen by the $\tt MCR$ method. It avoids any risk of colliding with the arch but has a large chance to miss the target. The right one shows the grasping configuration chosen by $\tt MaxSuccess$. It reasons about the uncertainty of the sugar box and the target pudding box together and comes up with a better solution which can reach the target accurately with low risk of collisions.}
\end{figure}

Fig. \ref{fig:experiment on real vision system} demonstrates the performance of different methods with real vision data. As the scenarios have no obvious risk-free picking path, $\tt OSP$ suffers from many collisions. $\tt MCR-MLC$ also has a relatively high number of collisions, which confirms that the distribution of object poses have to be considered instead of only the most likely pose. The true pose for an object may not be the top response of pose estimation. In every scenario, the methods $\tt MSE$ and $\tt MSG$ outperform $\tt MCR-G$ and $\tt MCR-E$, which are good at finding safe paths but have low success rate of reaching the target. The $\tt MCR$  methods tend to find conservative paths to avoid obstacles, sacrificing the reasoning about target poses. This observation is highlighted in the arch scenario (Fig. \ref{fig:experiment on real vision system}, bottom-right bar graph). Fig. \ref{fig:MCR vs MaxSuccess} shows the advantage of the proposed method over $\tt MCR$ in the arch scenarios.

\section{Discussion and Future Work}
\label{Discussion and Future Work}

This paper tackles the problem of picking a target item in the presence of multiple objects, where there is uncertainty for both obstructing objects and the target item. Perception and planning pipelines have been proposed to address this challenge. Both simulated and real-world experiments demonstrate effectiveness of the proposed framework, which minimizes collision probability while maximizing the probability of reaching the target.  An interesting direction to extend this work is safe object retraction, as well as potentially clearing an entire bin of objects with uncertain poses, where the robot also needs to decide the order with which to reach the target items.  A computational improvement can be achieved by reducing the overhead of collision checking as the number of pose hypotheses increases.  One heuristic is to delay the collision checking \cite{hauser2015lazy} before finding a plausible path and check afterwards to save computation.  An important extension is to deal with unknown objects or scenarios where object models are not available. In such cases a perception system can predict a volumetric representation for the objects' poses and shape uncertainty \cite{mitash2020task}, which can be incorporated into the proposed planning framework.

\bibliographystyle{format/IEEEtran}
\bibliography{bib/iros20}

\section*{Appendix}
\label{Appendix}

\subsection{Reduction from {\tt MCR} to stochastic {\tt MCR}}
The hardness of stochastic {\tt MCR} can be deduced via a reduction from {\tt MCR}, 
the
 hardness of which has been shown by polynomial-time reduction from SET-COVER \cite{hauser2014minimum}.

For {\tt MCR}, the input
corresponds to an initial state $q_s$, the goal state $q_g$ and $N$
objects $\mathcal{O}_{obj} = \{O_1, \dots ,
O_N\}$. 
For the
reduction, given any instance of {\tt MCR}, the input to a corresponding stochastic {\tt MCR} instance is generated according to the following polynomial-time process:

{\em (i)} the start $q_s$ and the goal $q_g$ remain the same as in {\tt MCR};
    
{\em (ii)} for each object $O_i$, the pose $P_i$ for object $O_i$
(deterministic) in {\tt MCR} defines a single pose
$p_i^1$ in stochastic {\tt MCR}. For the target, $p_t^1$ is the only target pose for stochastic {\tt MCR} and as a consequence, the target pose the goal configuration can pick is $J(q_g) = \mathcal{T}=[1]$. Here $q_g$ in {\tt MCR} is
equivalent to a single goal $g_g^1 \in \mathcal{G}$ in stochastic {\tt MCR};

{\em (iii)} all poses $p_i^1 (i=1, \cdots, N)$ are assigned the same
probability $\xi$, i.e., $Pr(p_i^1) = \xi \ (0 \leq \xi \leq 1)$.

Suppose the stochastic {\tt MCR} solver is able to find a solution path
$\pi^*$ of maximum $Succ(\pi^*)$ value (Eq. \ref{eq:max_success}) with $m$ poses intersected ($m \leq N$). 
In this case, $\mathds{1}_{\pi^*}(1,i) = 1$ for those $m$ intersected poses and $\mathds{1}_{\pi^*}(1,i) = 0$ for the remaining $N-m$ poses (Eq. \ref{indicator random variable}). 
Then according to Eq. \ref{survivability in labeled roadmap} where $j=1$, survivability $S_{\pi^*}$ is
\begin{equation}
    \label{survivability for pi_star}
    S_{\pi^*}
    = \prod_{i=1}^N 
    (1 - w(l_i^1)\mathds{1}_{\pi^*}(1,i))
    = (1 - \xi)^m
\end{equation}
For the term $Pr(q_g|\pi^*)$, there are two cases
\begin{enumerate}
    \item The path $\pi^*$ intersects the target pose $p_t^1$.
    
    In this case, $\hat{J}_{\pi^*}(q_g) = J(q_g) \setminus {\overline{J}}_{\pi^*} = \emptyset$, 
    which indicates that there are no remaining valid poses. 
    According to Eq. \ref{conditional definition of Pr(q_{g})}, 
    $Pr(q_g|\pi^*) = 0$. 
    Then 
    $Succ(\pi^*) = S_{\pi^*} \cdot Pr(q_{g}|\pi^*) = 0$,
    which indicates that the stochastic {\tt MCR} instance is not solvable. This also indicates no solution in original {\tt MCR} problem as the robot cannot grasp the target object $O_t$ while colliding with it.
    
    \item The path $\pi^*$ does not intersect the target pose $p_t^1$.
    
    In this case, $\hat{J}_{\pi^*}(q_g)$ remains the same since $p_t^1$ is still available and $Pr(q_g|\pi^*) = Pr(p_t^1) = \xi$. Then
    $Succ(\pi^*) = S_{\pi^*} \cdot Pr(q_{g}|\pi^*) = (1-\xi)^m \xi$. 
    To maximize the quantity $(1-\xi)^m \xi$, parameter $m$ needs to be
    minimized when $\xi$ is fixed, which corresponds to the objective of original {\tt MCR} problem (minimize the number of constraints violated).
\end{enumerate}

\begin{figure}[t]
    \centering \includegraphics[width = 0.48
      \textwidth]
      {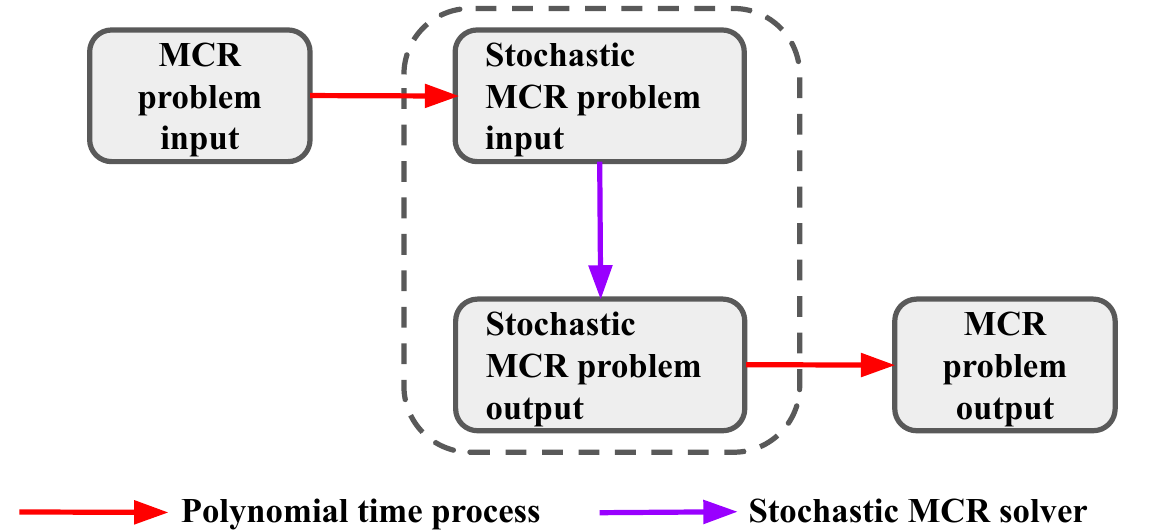}
    \vspace{-.1in}
\caption{\footnotesize 
    \label{fig:reduction diagram} 
    Diagram for the reduction from MCR to stochastic MCR}
\end{figure}

As a result, reconstructing the {\tt MCR}
solution from the stochastic {\tt MCR} one takes polynomial time.
Since the input/output transforms take polynomial time (Fig. \ref{fig:reduction diagram}), 
if the stochastic {\tt MCR} solver can return a solution in polynomial time, 
then {\tt MCR} is in P. 
Nevertheless, {\tt MCR} is NP-hard
\cite{hauser2014minimum}. 
Consequently, stochastic {\tt MCR} is an NP-hard problem.

\subsection{Roadmap generation}
The roadmap is generated as the input for the planning pipeline in this paper. Here the connectivity of the roadmap is defined similar to the ({\tt PRM}$^*$) variant, which achieves asymptotic
optimality \cite{Karaman:2011aa}, i.e., each node is connected to at
least $k^* = k_n \cdot log(n) = e(1+1/d)log(n)$ neighbors, where $e$
is the base of the natural logarithm, $d$ the dimension of the search
space (d=7) and $n$ the number of samples (n=5000).

\subsection{Accuracy of objects recognition}
As described in Section \ref{Perception pipeline}, one branch for the perception pipeline is to output the probability of each object detected in the scene. In the real-world experiment, the objects with probability $X_i$ larger than 0.3 are considered as existent in the scene. Table 1 shows the statistics of the object recognition process as a confusion matrix.

\vspace{0.04in}
\begin{table}[h!]
\centering
\begin{tabular}{|c|c|c|}
    \hline
     & Actually exist & Actually not exist \\
    \hline
     Predict as existent & 150 & 5 \\
     \hline
     Predict as not existent & 3 & 222 \\
     \hline
     \noalign{\vskip 1mm}
     \multicolumn{3}{c}{Table 1: Confusion matrix for object recognition.}
\end{tabular}
\label{table: confusion_matrix}
\end{table}

Based on Table 1, precision and recall are computed as
\begin{equation}
    \text{precision} = 150 / (150 + 5) = 96.8\%
\end{equation}
\begin{equation}
    \text{recall} = 150 / (150 + 3) = 98.0\%
\end{equation}

\vspace{0.06in}
The high precision ($96.8\%$) demonstrates the accuracy of excluding phantom objects (prevent extreme conservative plans) while the high recall ($98.0\%$) indicates the accuracy of detecting objects which are truly in the scene (critical for safe operation of the robot).

\subsection{Path cost and planning time}
The path cost (Euclidean distance between arm configurations along the path) and the planning time have also been computed for reference in Table 2, though they are not the main objectives considered in this paper. 

\vspace{0.04in}
\begin{table}[h!]
\centering
\begin{tabular}{|c|c|c|c|c|c|c|}
    \hline
     & OSP & MCR-G & MCR-E & MLC & MSG & MSE \\
    \hline
    cost & 3.023 & 3.627 & 3.705 & 3.255 & 4.221 & 4.425 \\
    \hline
    time(s) & 0.030 & 0.031 & 0.033 & 0.031 & 0.045 & 0.046 \\
    \hline
    \noalign{\vskip 1mm}
    \multicolumn{7}{c}{Table 2: Path cost and planning time.}
\end{tabular}
\label{table: path cost and planning time}
\end{table}

\end{document}